\documentclass[letterpaper, 10 pt, conference]{ieeeconf}  

\IEEEoverridecommandlockouts                              

\usepackage{graphics} 
\usepackage{epsfig} 
\usepackage{times} 
\usepackage{amsmath} 
\usepackage{amssymb}  
\usepackage{amsmath,amsfonts} 
\usepackage{url}
\usepackage{comment}
\usepackage{makecell, multirow}
\usepackage{multirow}
\usepackage{booktabs}
\usepackage{amsmath}
\usepackage{array}
\usepackage{hyperref}
\usepackage{mathrsfs}
\usepackage{amssymb}
\usepackage{amsfonts}
\usepackage{amsmath}
\usepackage{cleveref}
\usepackage[flushleft]{threeparttable}
\usepackage{tablefootnote}
\usepackage{multirow}
\usepackage{hhline}
\usepackage{graphicx}
\usepackage{graphicx, subcaption}
\usepackage{hanging}
\usepackage{color}
\usepackage{tikz}
\usepackage{float,wrapfig}
\usetikzlibrary{calc,arrows,decorations.markings}
\usepackage{balance}
\usepackage{hhline}
\usepackage{algorithm}
\usepackage{mathtools}
\usepackage{algpseudocode}
\usepackage{svg}
\usepackage{ifthen}
\usepackage{xspace}

\usepackage[normalem]{ulem}

\usepackage{etoolbox}

\usepackage[bottom]{footmisc}
\let\labelindent\relax
\usepackage[inline]{enumitem}
\setitemize{label=\textbullet, leftmargin=15pt, nolistsep}

\newcommand{\printfnsymbol}[1]{%
  \textsuperscript{\@fnsymbol{#1}}%
}

\title{Web2Grasp: Learning Functional Grasps from \\Web Images of Hand-Object Interactions}

\begin{document}
\author{Hongyi Chen$^{1*}$, Yunchao Yao$^{1*}$, Yufei Ye$^{2}$, Zhixuan Xu$^{3}$, Homanga Bharadhwaj$^{1}$, Jiashun Wang$^{1}$, \\ Arthur Jakobsson$^{1}$, Ruihan Zhao$^{4}$, Shubham Tulsiani$^{1}$, Zackory Erickson$^{1}$, and Jeffrey Ichnowski$^{1}$ \\
$^1$ Carnegie Mellon University, $^2$ Nvidia, $^{3}$ National University of Singapore, $^{4}$ UT Austin\\}
%

\newbool{DRAFT}
\setbool{DRAFT}{false}
\newcommand{\modelName}{\textbf{Web2Grasp}}
\newcommand{\myparagraph}[1]{\vspace{-5pt}\paragraph{#1}}
\newcommand{\expect}[2]{\mathbb{E}_{#1} \left[ #2 \right] }
\newcommand{\COMMENT}[3]{\ifbool{DRAFT}{\textcolor{#1}{[#2 -#3]}}{}}

\maketitle
\thispagestyle{empty}
\pagestyle{empty}

\begin{abstract}
Functional grasping is essential for enabling dexterous multi-finger robot hands to manipulate objects effectively. Prior work largely focuses on power grasps, which only involve holding an object, or relies on in-domain demonstrations for specific objects. 
We propose leveraging human grasp information extracted from web images, which capture natural and functional hand–object interactions (HOI).
Using a pretrained 3D reconstruction model, we recover 3D human HOI meshes from RGB images. To train on these noisy HOI data, we propose to use: 1) an interaction-centric model to learn the functional interaction pattern between hand and object, and 2) geometry-based filtering to remove the infeasible grasps and physical simulation to retain grasps who can resist disturbance. 
In IssacGym simulation, our model trained on reconstructed HOI grasps achieves a 75.8\,\% success rate on objects from the web dataset and generalizes to unseen objects, outperforming baseline methods in both grasp success and functional quality. In real-world experiments with the LEAP hand and Inspire hand, it attains a 77.5\,\% success rate across 12 objects, including challenging ones such as a syringe, spray bottle, knife, and tongs. 
\end{abstract}


\section{Introduction}
Humans grasp objects with intent, holding and operating them in ways that support natural use. We aim for robots to learn this form of \textit{functional grasping}, defined as grasping an object in a manner that enables its intended purpose during dexterous manipulation.
For example, operating a power drill requires the fingers to be precisely positioned on the trigger. While prior work has made substantial progress in dexterous grasping, many studies focus on power grasps that hold the object in place, often applying nearly identical grasps to objects that actually require diverse poses for effective use~\cite{wang2023dexgraspnet, li2023gendexgrasp, wei2024d}. These methods commonly rely on force-closure estimation or reinforcement learning approaches~\cite{zhang2025robustdexgrasp, wan2023unidexgrasp++}. While learning functional affordances from images and object shapes has been explored~\cite{li2024shapegrasp, shaw2023videodex, brahmbhatt2019contactgrasp}, studies have shown that learning grasp poses from such high-dimensional data remains challenging and often requires additional robot demonstrations to fine-tune the trained prior~\cite{shaw2023videodex, agarwal2023dexterous}. High-quality human-collected datasets that account for functional constraints have been developed~\cite{brahmbhatt2019contactdb, yang2022oakink, wei2024learning} and are used to train robots for functional grasping; however, curating such accurate contact data is costly.

To avoid time-consuming human contact collection across object categories, we explore an alternative source for acquiring functional grasp information: images of human hand-object interactions (HOI) from the web.
The images depict how humans grasp objects in functional ways, and recent advances in 3D vision for hand-object pose estimation and reconstruction from RGB images~\cite{pavlakos2024reconstructing, wen2024foundationpose} can enable extracting this information. 
However, due to mutual occlusions and the lack of annotated training data, these vision models are often inaccurate, producing reconstructed HOI that exhibit improper contact with object meshes~\cite{ye2022s, ye2023affordance}, making them difficult to use for accurate robot grasping. Some prior works leverage reconstructed 3D data in simulators~\cite{singh2024hand, torne2024reconciling} to improve grasping and manipulation, but they either use parallel grippers or focus on power grasps, leaving functional multi-fingered grasping underexplored.

\begin{figure}[t]
  \begin{center}
    \includegraphics[width=0.9\linewidth]{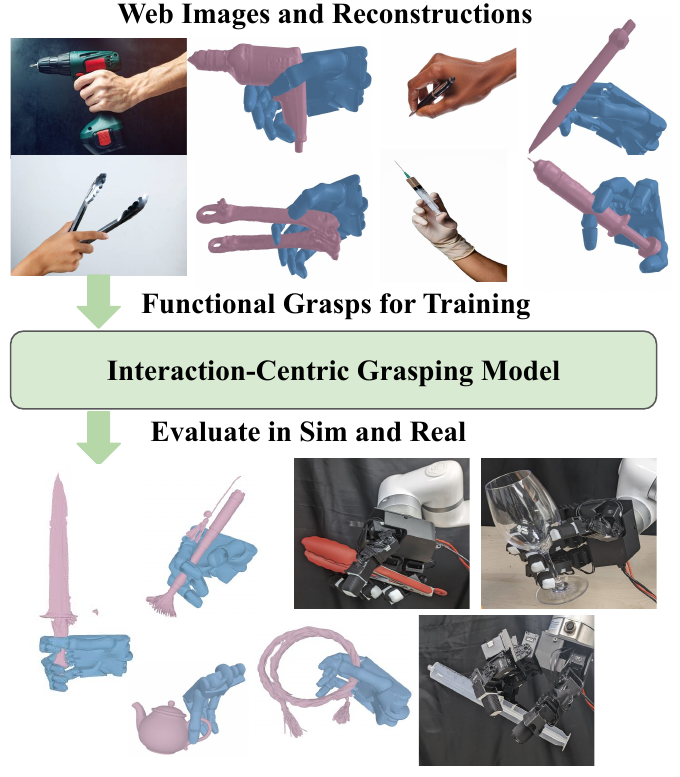}
  \end{center}
  \vspace{-10pt}
  \caption{\small \textbf{Top:} Web images and reconstructed hand-object interaction (HOI) grasps. \textbf{Bottom:} Functionally valid grasps in simulation and in the real-world.}
  \vspace{-10pt}
  \label{fig:teaser}
\end{figure}

We propose \modelName, an approach to enable learning functional grasps using reconstructed HOI data from web images as a rich source of functional object interaction. 
Our method involves reconstructing HOI data and exploiting this noisy data to train a functional grasping model. Specifically, we obtain HOI data from web-crawled images of humans interacting with objects using a pretrained 3D reconstruction model~\cite{ye2022s}, and align the low-quality object mesh reconstructions with 3D shapes generated by the text-to-3D tool \emph{MeshyAI}~\cite{meshyai}. 
Since these reconstructed object meshes are often noisy and can lead to unrealistic finger contacts, we propose: 1) an interaction-centric model that learns functional interaction patterns between hand and object, rather than directly predicting robot hand joints, and 2) geometry-based filtering to remove infeasible grasps, combined with physical simulation to retain grasps that can resist disturbances and expand the dataset. We evaluate Web2Grasp on both simulation and real-world functional grasps across diverse objects using different dexterous hand embodiments. Compared to baseline methods, our grasping model trained on reconstructed HOI data achieves high success rates as well as superior functionality, measured by both the functional region contact ratio and human evaluation.

\noindent \textbf{Contributions.}
\begin{enumerate*}[label=(\roman*), itemjoin=\quad]
    \item A new dexterous grasping training method, \modelName, that leverages reconstructed HOI from web images. 
    \item Interaction-centric modeling combined with data-wise refinement to effectively exploit reconstructed HOI for dexterous grasp training.
    \item Comprehensive evaluations in both simulation and the real world, demonstrating that the grasping model trained on HOI data from web images achieves high success rates while preserving human-like functional grasping patterns.
\end{enumerate*}


\section{Related Work}
\subsection{Dexterous Robotic Grasping and Functional Grasping}
Dexterous grasping is essential for enabling robots to manipulate objects with greater versatility and agility compared to parallel grippers~\cite{fang2023anygrasp, bialek2023comparative, deng2025graspvla}. However, the high dimensionality of dexterous hands makes grasp synthesis significantly more challenging than parallel-jaw or suction grippers. Classical approaches formulate grasping as an optimization problem that maximize force closure objectives~\cite{wang2023dexgraspnet, li2023gendexgrasp, brahmbhatt2019contactgrasp}. Some methods leverage contact information between the hand and object to learn effective grasps~\cite{brahmbhatt2019contactgrasp}, while others employ reinforcement learning to acquire grasping strategies in simulation~\cite{wan2023unidexgrasp++, qin2023dexpoint}. Although simulation can optimize grasp metrics, much existing work focuses on power grasps for stable holding rather than functional grasps that support complex in-hand manipulation~\cite{wang2023dexgraspnet, li2023gendexgrasp}.

To learn feasible and \emph{functional} grasps, some approaches rely on clean annotations or human-collected grasp datasets~\cite{brahmbhatt2019contactdb, yang2022oakink, taheri2020grab}. Other works aim to learn affordances directly from human images or videos~\cite{agarwal2023dexterous, bahl2023affordances}, either via retargeting or reinforcement learning. However, these functional grasping methods are typically evaluated using parallel grippers or on geometrically simple objects. A separate line of research explores the use of large language models (LLMs) to infer functional contact points and enable zero-shot grasping~\cite{li2024shapegrasp, xu2023creative}. However, these methods have not yet been demonstrated on dexterous hands. Overall, visual data like images are noisy and difficult to learn from directly, frequently requiring human demonstrations to ensure grasp quality~\cite{shaw2023videodex, agarwal2023dexterous}. In contrast, we first reconstruct the HOI mesh from images, where contact information is available and well-suited for dexterous grasp learning.  


\subsection{Human-Object Interaction from Images and Videos.} Hand pose estimation from RGB(\nobreakdash-D) images has seen significant advancements, particularly through statistical models such as MANO~\cite{MANO:SIGGRAPHASIA:2017}, whose low-dimensional pose and shape parameters can now be directly regressed or learned with modern techniques like transformers~\cite{rong2020frankmocap}. While reconstructing objects from images remains a challenging task, recent developments have demonstrated that, with strong 3D supervision, it is possible to learn a shared model across multiple object categories. These models output representations in various formats, such as voxels~\cite{choy20163d}, meshes~\cite{gkioxari2019mesh}, point clouds~\cite{lin2018learning}, and neural-implicit~\cite{park2019deepsdf, ye2022s}. With these tools, joint reasoning of hand-object interactions (HOI) from images and videos has become increasingly feasible, particularly through 3D learning models~\cite{ye2022s, wu2024reconstructing, fan2024hold}. Building on these works, researchers have started leveraging reconstructed HOI data in simulation environments to train robot grasp and manipulation policies~\cite{wang2023real2sim2real}, often by transferring human hand motions to robot hands~\cite{huang2025fungrasp}. The closely related work~\cite{singh2024hand} focuses on learning manipulation priors from reconstructed HOI and requires downstream fine-tuning, but it does not fully exploit HOI interactions for functional grasps. In contrast, our work demonstrates that effective functional grasping models can be trained using reconstructed HOI data, despite its typically lower quality compared to human-collected datasets.

\section{Method} 
\label{sec:method}

Given point cloud observations $\mathbf{P}^{\mathcal{O}}$ of diverse objects, our goal is to learn a grasping model that can generate functional grasps $q$ that mimic human interaction patterns, assuming that humans grasp objects with intent to support their natural use.
We introduce \textbf{Web2Grasp}, a method that (i) reconstructs human grasps from web-scraped RGB images and transfers them to robotic hands (Sec.~\ref{sec:recon}); (ii) trains an interaction-centric grasping model via supervised learning to imitate the functional grasping behaviors observed in the images (Sec.~\ref{sec:dro}); and (iii) grasps filtering and evaluating in simulation to ensure they can withstand force disturbances (Sec.~\ref{sec:sim_aug}).

\subsection{Reconstructing Robotic Hand-Object Interaction}
\label{sec:recon}
\begin{figure*}[t]
    \centering
    \includegraphics[width=0.85\linewidth]{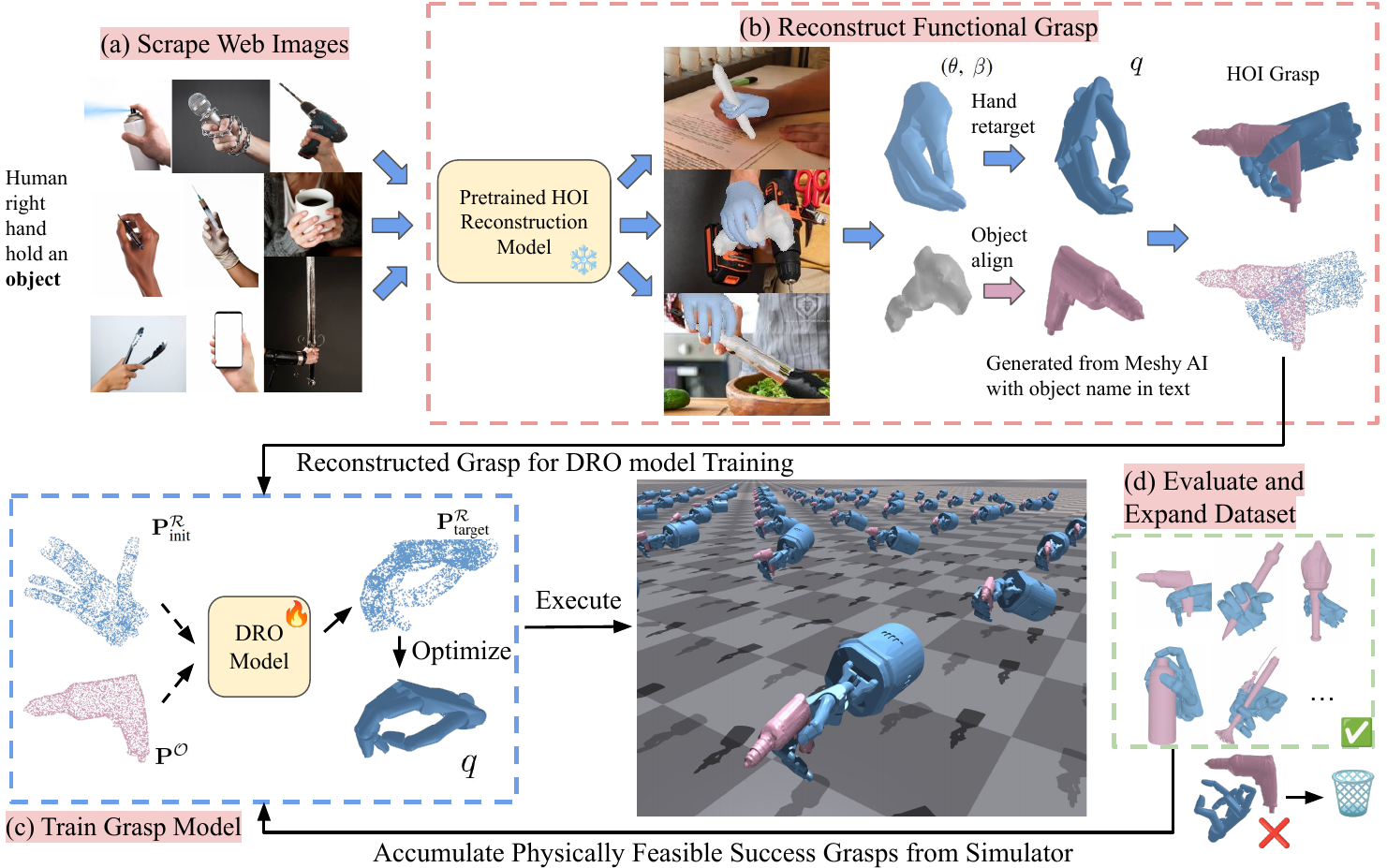}
    \caption{\small \textbf{Hand-Object Interaction (HOI) Collection, Grasp Model Training and Execution Pipeline.} \modelName (a) collects images of humans grasping objects from the web, the (b) uses HOI reconstruction to produce a functional grasp dataset, potentially containing penetrations and unrealistic contacts. (c): Web2Grasp trains a DRO grasping model on the HOI dataset to predict target joint configurations for grasp execution. (d): Web2Grasp uses simulation to collect physically feasible grasps to expand the dataset and retrain the model. 
    } 
    \vspace{-5pt}
    \label{fig:pipeline}
\end{figure*}

Web2Grasp begins by crawling web images of humans holding various objects. 
Given a randomly scraped image $I$ showing a human hand holding an object from a specific category, the goal is to recover the corresponding robot hand configuration and the object’s interaction model, which will later be used for training a robot grasping model. Specifically, we adopt the method proposed by Ye et al.~\cite{ye2022s} to predict the MANO hand pose ($\theta$, $\beta$), a low-dimensional representation of the human hand, using the off-the-shelf FrankMocap~\cite{rong2020frankmocap}, and infer the shape of the interacting object in the same frame via a signed-distance function (SDF) conditioned on the hand pose. The reconstructed object meshes are often of low quality, with imprecise geometry and visually unrealistic appearances (e.g., the gray power drill in Fig.~\ref{fig:pipeline}~(b)), particularly in cases of significant hand occlusion or when the objects are outside the training distribution~\cite{ye2022s, wu2024reconstructing}. Thus, we align it with an accurate 3D mesh from the same category using the Iterative Closest Point (ICP) algorithm. We obtain 3D meshes by inputting the object name as text into online mesh generators \emph{MeshyAI}~\cite{meshyai} and \emph{Genie}~\cite{lumagenie}. Our goal is not to achieve pixel-perfect object reconstruction, but to preserve key interaction cues that enable a robot hand to interact naturally and functionally, similar to human behavior, as shown in Fig.~\ref{fig:pipeline}~(a) and (b). Moreover, our method is agnostic to the choice of mesh reconstruction and can benefit from improvements in reconstruction techniques.

To leverage the reconstructed human hand-object interactions for robot grasping, we transfer the MANO hand poses to the robot hand configurations $q$. We employ position-based optimization to minimize the 3D position error between keypoints on the robot's links and their corresponding points on the MANO hand~\cite{qin2023anyteleop}. The final robot HOI representation consists of the object mesh centered at the origin, while the robot hand configuration $q$ encodes the wrist’s relative position and orientation with respect to the object, along with the finger joint configuration.

\subsection{Interaction-Centric Grasping Model Training} 
\label{sec:dro}
While reconstructed robot HOI data from web images capture functional interactions, robot grasps obtained through direct retargeting are often physically implausible due to the embodiment gap between human and robot hands, inaccuracies in hand detection, and mismatches between web-image objects and aligned target objects.
To leverage this dataset for training a model capable of generating grasps across diverse objects, we address the challenge from two perspectives: (1) algorithmically, by employing a more robust model to handle noise in HOI grasps (discussed in this section), and (2) data-wise, by filtering out poorly reconstructed grasps through geometry-based and simulation-based evaluation (Sec.~\ref{sec:sim_aug}).

Instead of relying on robot-centric grasping models~\cite{liu2021synthesizing, wan2023unidexgrasp++}, which require accurate target joint configurations, or object-centric models~\cite{li2023gendexgrasp}, which depend on precise contact maps of object shapes, we adopt a more robust alternative: the interaction-centric model DRO~\cite{wei2024d}. DRO predicts dense point-to-point distances between the robot hand in a grasp configuration and the object, making it well-suited for learning functional grasp patterns from imperfect HOI data. We therefore use DRO to train on our reconstructed dataset.

Given the initial robot point clouds $\mathbf{P}_{\text{init}}^{\mathcal{R}}\in \mathbb{R}^{N_{\mathcal{R}} \times 3}$ and object point clouds $\mathbf{P}^{\mathcal{O}}\in \mathbb{R}^{N_{\mathcal{O}} \times 3}$ sampled from the robot and object mesh in HOI data, the objective of the DRO model is to predict the point-to-point distance matrix $\mathcal{D(R, O)}^{\text{Pred}} \in \mathbb{R}^{N_{\mathcal{R}} \times N_{\mathcal{O}}}$, shown in Fig.~\ref{fig:pipeline} (c). The DRO model uses a Conditional Variational Autoencoder (CVAE)~\cite{sohn2015learning} to capture features across diverse hand-object interactions and to predict pairwise relative distances, following the design in Eisner et al.~\cite{eisner2024deep}. The training loss is the difference between the predicted and ground-truth distance matrices with $\mathcal{L_{\text{L1}}}\left( \mathcal{D(R,O)}^{\text{Pred}}, \mathcal{D(R,O)}^{\text{GT}} \right)$. Using $\mathcal{D(R,O)}^{\text{Pred}}$ and the object point clouds $\mathbf{P}^{\mathcal{O}}$, we can position the robot point cloud in the target grasp pose $\mathbf{P}_{\text{target}}^{\mathcal{R}}$ with a multilateration method~\cite{norrdine2012algebraic} and compute the joint configuration through optimization. See Wei et al.~\cite{wei2024d} for details.

\subsection{Reconstructed HOI Dataset Refinement} 
\label{sec:sim_aug}
This section discusses data filtering to enhance the quality of the reconstructed dataset for training the DRO grasping model. As noted above, many samples contain hand-object interpenetration or improper contact. Training directly on such data often leads the model to predict grasps with incorrect contact locations and unstable configurations. To address this, we employ the following filtering process:
\begin{enumerate}
    \item Geometry-based filtering: applying thresholds on low penetration, low distance and high contact ratio between hand and object meshes to remove infeasible HOI grasps. This step reduces the full web-reconstructed dataset to a smaller subset without requiring simulation.
    \item Simulation-based evaluation: We first train an initial DRO model on the filtered dataset. The model is then used to generate a large number of grasps for evaluation in physical simulation, where external force disturbances are applied and grasps that fail to securely hold objects are discarded. This process both expands the dataset and improves its robustness.  
\end{enumerate}

Combining all of the ingredients above, we propose a four-step pipeline to collect functional training grasps from web images and improve our model: 
\begin{enumerate*}[label=(\arabic*)] 
    \item Perform HOI reconstruction from web images and apply geometry-based filtering; 
    \item Train the DRO model on the filtered reconstructed grasps; 
    \item Use the learned model to generate more grasps in simulation and keep those that resist force disturbances to expanding the dataset;
    \item Retrain the model using the grasps that pass both the filtering and simulation evaluation.
\end{enumerate*}

Notably, Steps (3) and (4) are executed sequentially as a one-time data augmentation and retraining procedure, rather than iteratively. This pipeline generates functional grasps that reflect human-intended object use, by training the model on interaction patterns reconstructed from web images and expanding the dataset with simulation-augmented grasps that withstand external disturbances. In this way, the final dataset and the derived model are both functionally and physically plausible. In Sec.~\ref{sec:sim_grasp}, we further demonstrate this empirically.


\section{Experiments}

We evaluate Web2Grasp across three experimental settings: (1) the quality of reconstructed HOI data, (2) the success rates and functionality of predicted grasps in simulation using ShadowHand, and (3) real-world grasping experiments with LEAP hand and Inspire hand.

\subsection{HOI Reconstruction Results}
\label{sec:exp_sim}

\begin{figure*}[t]
    \centering
    \includegraphics[width=0.9\linewidth]{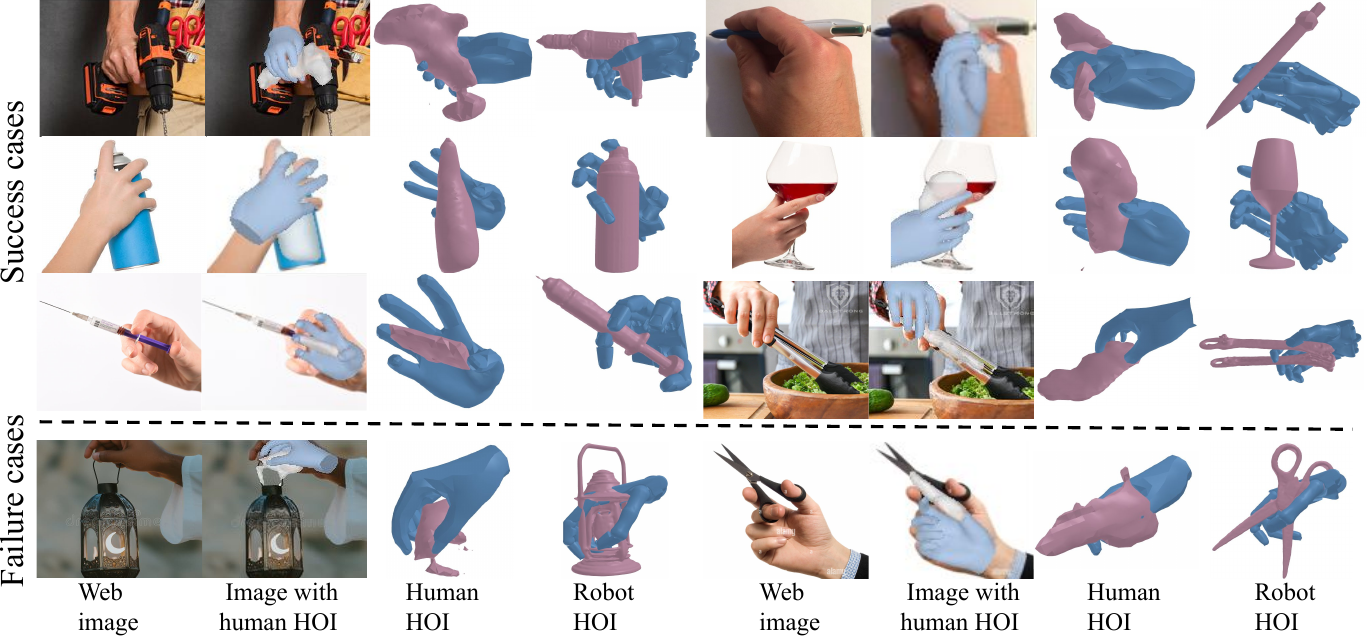}
    \vspace{-5pt}
    \caption{\small
        \textbf{Visualization of HOI Reconstruction from Web Images.} Reconstructions for representative objects. \textbf{Success cases} include Row 1: \emph{Power Drill}, \emph{Pen}; Row 2: \emph{Spray Bottle}, \emph{Wine Glass}; Row 3: \emph{Syringe}, \emph{Tongs}. \textbf{Failure cases} include Row 4: \emph{Lantern}, \emph{Scissors}. From left to right: original web image, web image overlaid with human HOI mesh, human HOI mesh alone, and robot HOI mesh with retargeted robot hand and aligned object.
    }
    \label{fig:grasp_comparison}
    \vspace{-10pt}
\end{figure*}

\textbf{Setup and Baselines} Our reconstructed HOI dataset includes 10 object categories, each with 100 reconstructions from randomly scraped web images. 
We evaluate the grasp quality of HOI dataset and compare it with the synthesized grasps from the \emph{DRO Dataset} used to train the DRO model~\cite{wei2024d}, as well as the human-collected OakInk grasp dataset~\cite{yang2022oakink}, using the following metrics~\cite{yang2022oakink, jiang2021hand}: (1) \emph{Penetration Score}, quantifies the penetration depth and volume between the hand and the object; (2) \emph{Disjoint Distance}, computes the average distance from hand fingers to the nearest object surface, contrast to penetration; (3) \emph{Contact Region Ratio}, measures the proportion of surface points on the object in contact with the robot hand. We provide both visual and numerical results in Fig.~\ref{fig:grasp_comparison} and Tab.~\ref{tab:grasp_qua}. 

\begin{table}[t]
\caption{\small \textbf{Comparison of grasp quality metrics across datasets.} Lower is better for penetration and disjoint distance, while higher contact ratio indicates better contact coverage if penetration is low.} 
\vspace{-10pt}
\begin{center}
\small
\scalebox{0.83}{
\begin{tabular}{@{}cccc@{}}
\toprule
& Penetration Depth [cm] & Grasp Disjoint & Contact \\
& (Volume [cm$^3$]) $\downarrow$ & Mean [cm] $\downarrow$ & Ratio [\%] $\uparrow$ \\
\midrule
DRO dataset~\cite{wei2024d} & 0.14 (1.44) & 0.26 & \phantom{1}5.81 \\
OakInk dataset~\cite{yang2022oakink} & 0.11 (0.62) & 0.09 & 10.97 \\
Full web dataset & 1.11 (9.79) & 0.18 & \textbf{11.36} \\
Filtered web dataset & 0.59 (4.85) & \textbf{0.08} & \phantom{1}9.32 \\
Sim dataset & \textbf{0.01} (\textbf{0.13}) & 0.14 & 1.44 \\
\bottomrule
\end{tabular}
}
\label{tab:grasp_qua}
\end{center}
\vspace{-10pt}
\end{table}

\textbf{Results} 
In the quantitative evaluation (Tab.~\ref{tab:grasp_qua}), our dataset exhibits high penetration scores, lower disjoint distances and higher contact ratios compared to the synthesized DRO dataset, where frequent interpenetration can artificially inflate these metrics. The human-collected OakInk dataset has the best overall balance, with the lowest penetration and the second-highest contact ratio. This highlights the importance of geometry-based filtering to retain grasps with low penetration, low disjoint distance and high contact ratios. The filtering process yields a subset of 96 grasps, representing 9.6\% of the full web dataset. We use this subset for training and refer to it as \emph{web data} in the following sections (Sec.~\ref{sec:sim_grasp}). To augment this relatively small dataset, we further collect successful grasps in IsaacGym under force disturbances to construct a simulation dataset (\emph{sim data}) comprising 3,800 grasps (details in Sec.~\ref{sec:sim_grasp}). These simulated grasps exhibit minimal penetration, as the physical engine inherently prevents deep interpenetration, while maintaining small distance between the fingers and the object. Compared to the web data, the simulation data shows a lower contact ratio, since penetration in the web data artificially inflates it. After removing penetration, the separation between the hand and object reveals a more realistic contact ratio.

Qualitatively, many reconstructed grasps preserve the functional patterns observed in the original web images (first three rows of Fig.~\ref{fig:grasp_comparison}). However, HOI reconstruction struggles with objects of complex geometry or interaction modes, such as buckets and scissors. For buckets, thin structures (e.g., handles) are often missed, causing the aligned robot hand to grasp the body instead of the handle. Scissors are challenging due to their unique interaction mode, where fingers are inserted into handle holes (last row of Fig.~\ref{fig:grasp_comparison}).

\subsection{Simulation Grasp Experiment Setup}
\label{sec:sim_grasp}

\begin{table*}
\caption{\small \textbf{Grasp Success Rates in IsaacGym for Models Trained on Web Dataset and Simulation Augmented Dataset.} Objects from \emph{Power Drill} to \emph{Sword} are seen during training, while \emph{Whip} to \emph{Writing Brush} are unseen. Both methods are evaluated over 100 trials per object.}
\centering
\small
\scalebox{0.9}{
\begin{tabular}{lcccccccccc}
\toprule
Dataset & \begin{tabular}[c]{@{}c@{}}Power\\ drill\end{tabular} & Pen & Microphone & Phone & \begin{tabular}[c]{@{}c@{}}Spray\\ bottle\end{tabular} & \begin{tabular}[c]{@{}c@{}}Wine\\ glass\end{tabular} & Tong & Syringe & Mug & Sword \\
\midrule
Web data & 98 & 92 & 98 & 62 & 74 & 86 & 72 & 64 & 88 & 24 \\ 
Sim data & 92 & 97 & 99 & 80 & 71 & 92 & 94 & 72 & 99 & 55 \\
\midrule
         & Whip & Teapot & Axe & Remote & Torch & Hammer & Whisk & \begin{tabular}[c]{@{}c@{}}Soap\\ bottle\end{tabular} & \begin{tabular}[c]{@{}c@{}}Writing\\ brush\end{tabular} & Average \\
\midrule
Web data & \phantom{0}8 & 12 & 10 & 92 & 82 & 94 & 20 & 80 & 19 & 61.8 \\ 
Sim data & 30 & 99 & 82 & 89 & 96 & 78 & 100 & 82 & 78 & \textbf{83.4} \\
\bottomrule
\end{tabular}
}
\label{tab:grasp_sim}
\end{table*}

Our method adopts the DRO architecture and train it with the filtered HOI dataset (\emph{web data}) described above. The training set contains grasps from 10 object categories: \emph{Power Drill}, \emph{Pen}, \emph{Microphone}, \emph{Phone}, \emph{Spray Bottle}, \emph{Wine Glass}, \emph{Tong}, \emph{Syringe}, \emph{Mug}, and \emph{Sword}. 
We evaluate on 9 test object categories with different geometries but similar functional grasp patterns: \emph{Whip}, \emph{Teapot}, \emph{Axe}, \emph{Remote}, \emph{Torch}, \emph{Hammer}, \emph{Whisk}, \emph{Soap Bottle}, and \emph{Writing Brush}. For instance, \emph{Spray Bottle} (train) and \emph{Soap Bottle} (test) share a similar top trigger functional grasp.

We evaluate both grasp success rate and grasp functionality using the ShadowHand in the IsaacGym simulator.
\begin{enumerate*}[itemjoin=\quad]
\item A grasp is considered successful if the object displacement remains below 2\,cm from its initial pose under external force disturbances. The disturbance magnitude is set to 0.5~times the object mass. Each grasp is subjected to a 600-step disturbance phase, during which forces are applied sequentially along six directions: $\pm x$, $\pm y$, and $\pm z$.
\item We assess grasp functionality through two complementary evaluations: functional region contact, which measures contact on affordance-critical areas required for object operation, and a human study that evaluates overall grasp quality.
\begin{enumerate*}[label=(\roman*), itemjoin=\quad]
\item We annotate functional regions on each object, such as the trigger on the \emph{Power Drill}, and the spray heads on the \emph{Spray}. For each predicted grasp, an object vertex is considered in contact if its distance to any hand finger vertex is less than 0.3\,cm. We compute the functional contact ratio (FCR) as $\text{FCR} = \frac{|V_{\text{contact}} \cap V_{\text{functional}}|}{|V_{\text{functional}}|}$, where $V_{\text{contact}}$ and $V_{\text{functional}}$ denote the contact and annotated functional vertex sets, respectively.
\item In the human study, 24 participants aged 19-52, from technical and non-technical backgrounds, evaluate overall grasp functionality and object interaction. Each participant compares 10 randomly selected method pairs and selects the grasp they find most suitable for the object’s intended use. We normalize the votes to compute functionality scores.
\end{enumerate*}
\end{enumerate*}

We compare to the following baselines:
(1)~\textit{GenDexGrasp}~\cite{li2023gendexgrasp}: Generates grasp poses for novel objects by leveraging contact maps as a hand-agnostic intermediate representation;
(2)~\textit{DexGraspNet}~\cite{wang2023dexgraspnet}: A model trained on a large-scale dexterous grasp dataset;
(3)~\textit{DRO}~\cite{wei2024d}: An interaction-centric model that predicts the relative distance between robot hand and object point clouds.
Through the experiments, we aim to answer the following questions:
\begin{itemize}
\item \textbf{Q1}: Can reconstructed web data effectively train a dexterous grasping model?
\item \textbf{Q2}: How does our model, trained on seen objects, perform compared to baselines in success rate and functionality on unseen objects with novel geometries?
\item \textbf{Q3}: Does augmenting the simulation dataset improve success rate while preserving grasp functionality?
\end{itemize}

\begin{figure*}[t]
    \centering
    \includegraphics[width=0.99\linewidth]{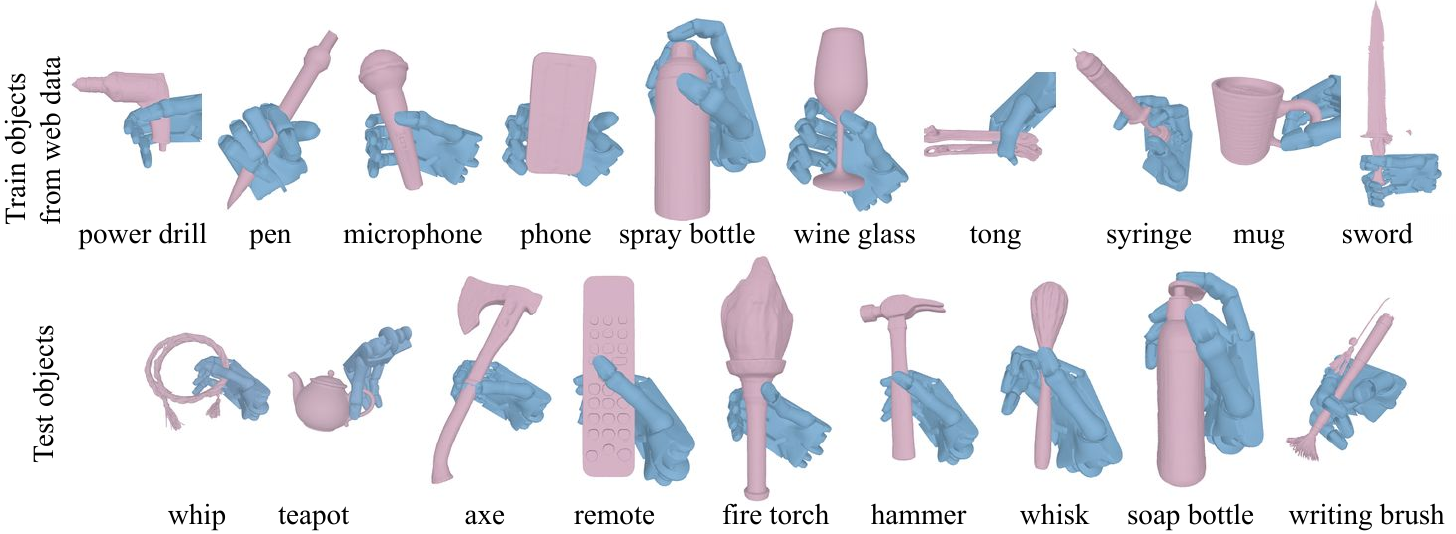}
    \caption{\small
        \textbf{Visualization of Generated Grasps Across All Objects.} The first row shows objects seen during web-data training, while the second row presents unseen objects from web images.
    }
    \label{fig:sim_grasp}
    \vspace{-10pt}
\end{figure*}

\subsection{Simulation Grasp Experiment Results}

\textbf{A1:} Our approach, trained on web data with 10 objects, achieves a 75.8\% success rate, as shown in the first row (web data) of Tab.~\ref{tab:grasp_sim}. This demonstrates that functional grasping models can be effectively trained using 3D reconstructed HOI from web data without additional robot demonstrations. The grasps also preserve human-like functional behaviors, as shown in Fig.~\ref{fig:sim_grasp}. We observe a lower success rate on the \emph{Sword}, as its grasps are easily disrupted by disturbances due to the lack of stabilizing force. But we show later that incorporating additional successful grasps from the simulator mitigates this issue.

\begin{figure}[t]
  \begin{center}
    \includegraphics[width=0.99\linewidth]{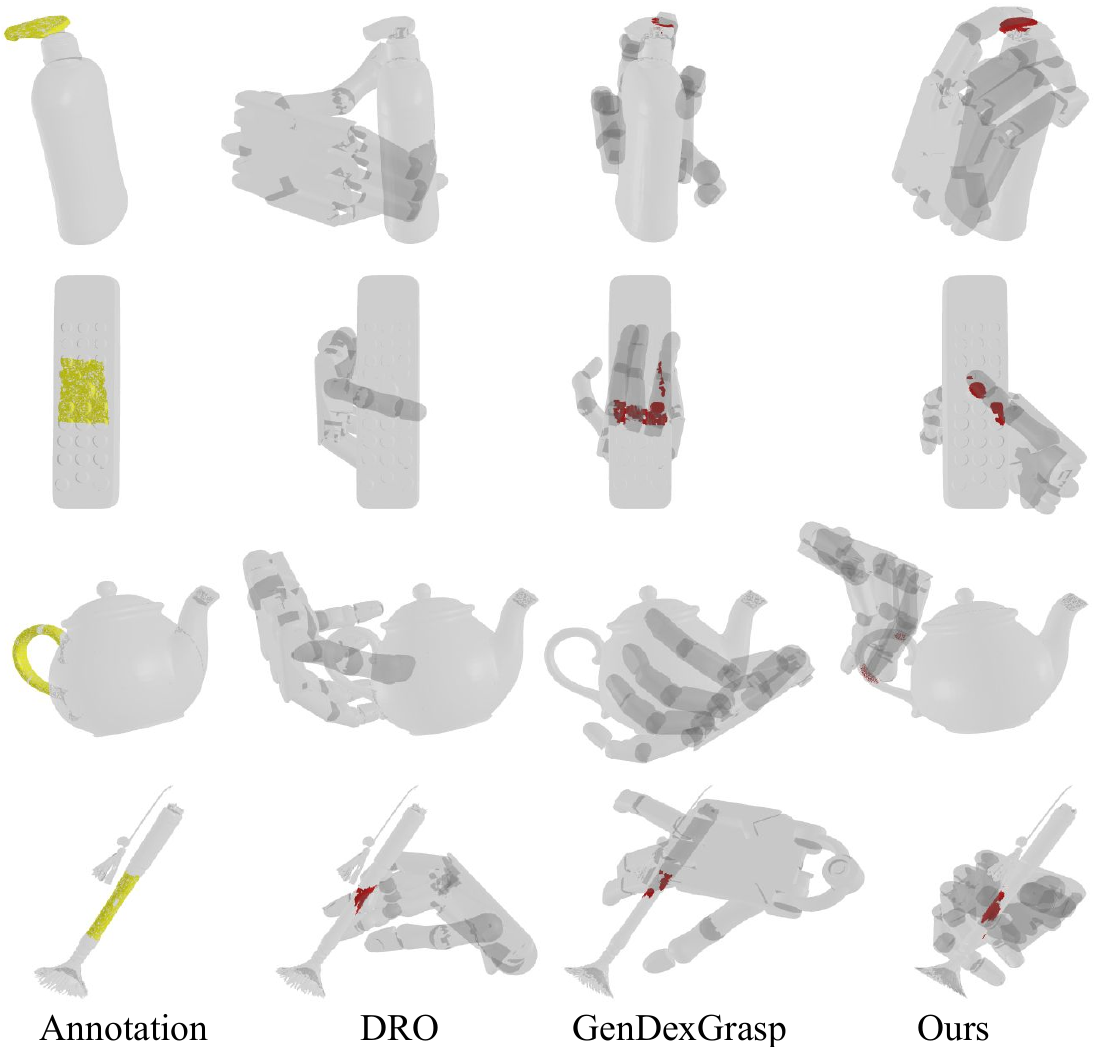}
  \end{center}
  \vspace{-5pt}
    \caption{\small \textbf{Annotated Functional Region and Contact Regions from Predicted Grasps Across Methods.} 
    The first column shows the manually annotated functional regions (yellow) on the \emph{Soap Bottle}, \emph{Remote}, \emph{Teapot}, and \emph{Writing Brush}. In the remaining columns, red denotes the contact regions between the fingers and the functional region for grasps predicted by different methods on this object.}
  \vspace{-15pt}
  \label{fig:fcr}
\end{figure}

\textbf{A2:} We train our model on web data and evaluate it, along with all baselines, on the same set of 9 unseen test objects. Our method outperforms baselines on these unseen objects in both success rate and functionality score (Tab.~\ref{tab:baseline_sim}), demonstrating that it learns nuanced interactions from web data and generalizes to unseen but related object categories, such as transferring from \emph{Pen} to \emph{Writing Brush}, from \emph{Mug} to \emph{Teapot}, and from \emph{Spray Bottle} to \emph{Soap Bottle}.
Baseline methods achieve high success rates on common objects that primarily require stable grasps but often fail to capture functional strategies. For example, DRO uses the same model architecture as ours but is trained on its original dataset~\cite{wei2024d}. It grasps the \emph{Soap Bottle} from the side instead of placing a finger on top to enable operation, resulting in a low FCR (Fig.~\ref{fig:fcr}). Although DRO reports improved dexterous grasping performance in its original paper, it does not learn human-like functional grasps because its training data are synthetically generated and lack functional supervision.
GenDexGrasp achieves a high FCR; however, its predicted grasps frequently penetrate the object surface, artificially reducing finger-object distances and inflating the FCR metric, which leads to a low human evaluation score. In contrast, our method achieves both the highest FCR and human evaluation score, indicating that the predicted grasps are plausible from a human perspective and functionally appropriate.

\begin{table}
\caption{\small \textbf{Grasping Performance of Unseen Objects in Simulation Across All Models.} 
SR denotes Success Rate. FCR denotes Functional Contact Ratio, defined as the percentage of contacted functional area relative to the total functional area.}
\centering
\begin{tabular}{lccc}
    \toprule
    & \multirow{2}{*}{SR (\%)} & \multicolumn{2}{c}{Functionality} \\
    &  & FCR (\%) & Human Eval \\
    \midrule
    GenDexGrasp~\cite{li2023gendexgrasp} & 20.6 & 6.81 & 6.4 \\
    DexGraspNet~\cite{wang2023dexgraspnet} & 39.6 & 3.58  & 19.6 \\ 
    DRO~\cite{wei2024d} & 33.4 & 3.53  & 26.3 \\ 
    Ours (Web data) & \textbf{46.3} & \textbf{7.91} & \textbf{47.7} \\
    \bottomrule
\end{tabular}
\vspace{-5pt}
\label{tab:baseline_sim}
\end{table}

\textbf{A3:} Since the reconstructed web dataset is relatively small compared to simulator-collected grasp datasets~\cite{wang2023dexgraspnet}, we deploy the web-trained model in Isaac Gym to collect 200 successful grasps per object, creating a simulation-augmented dataset (sim data) for retraining. The purpose of this experiment is not to evaluate generalization, but to examine whether simulation-based augmentation can improve the performance of the initial web-trained model. For this purpose, we include the original 9 test objects as well.
Web2Grasp, trained with web data and simulation-augmented datasets, achieves success rates of 61.8\,\% and 83.4\,\% across all 19 objects, respectively, as summarized in Tab.~\ref{tab:grasp_sim}.  
This demonstrates that even for certain objects in seen set (e.g., \emph{Sword}) or objects without web-image HOI reconstructions (e.g., \emph{Teapot}, \emph{Writing Brush}), as long as the web-trained model can generate a few successful grasps, these can be expanded into a larger training set to significantly boost success rates.
For instance, in the web dataset we observe difficulty in securely holding the \emph{Sword}. The simulation-augmented dataset mitigates these issues by saving only stable contacts that withstand external force disturbances. 
Moreover, the FCR of grasps generated by models trained with web data and simulation-augmented data is 8.37\% and 7.42\% respectively, indicating that the simulation refinement preserves the functional grasping behaviors learned from web data while improving stability.

\subsection{Real World Experiment}
\label{sec:exp_real}
\textbf{Setup.} We evaluate our method in two real-world settings using the dexterous LEAP hand~\cite{shaw2023leaphand} and the Inspire hand~\cite{inspire_hand}, each mounted on a 7-DoF robot arm. The goal is to evaluate our method’s ability to grasp real-world objects using web-image-based HOI data from the target category, without relying on demonstrations specific to the test objects. We reconstruct HOI data from web images, retarget it to the LEAP or Inspire hand for simulation training, and transfer the learned grasping models to the real world. For testing objects in real-wolrd, we either 3D-print these objects at the same scale as those used in HOI reconstruction or source real-world counterparts (e.g., \emph{Tong}, \emph{Syringe}) with closely matching sizes. This allows us to use known object meshes and sample point clouds for predicting grasps in real-world experiments. 
For each robotic hand, we train a dedicated grasping model that outputs 22 DoF (16 finger DoF and 6 end-effector DoF) for the LEAP hand, and 12 DoF (6 finger DoF and 6 end-effector DoF) for the Inspire hand, enabling object grasps. Our focus is on hand grasping, and design the arm movement trajectory to guide the hand to the object for both our method and baselines. To perform a grasp, we move the hand to a starting position, using the predicted wrist orientation while keeping the fingers open. The robot approaches the object and then closes fingers around the object according to the predicted finger configuration. The robot estimates the object pose using FoundationPose~\cite{wen2024foundationpose}. A successful grasp lifts and holds the object against gravity.

\begin{figure}[t]
  \begin{center}
\includegraphics[width=0.99\linewidth]{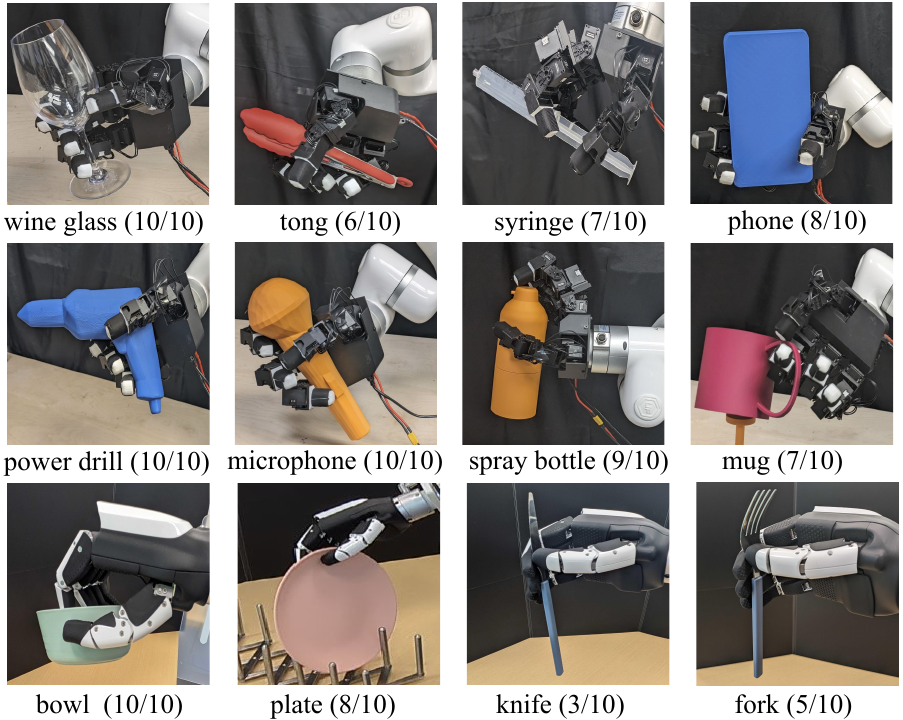}
  \end{center}
  \vspace{-5pt}
    \caption{\small \textbf{Visualization of Grasps and Success Rates with LEAP Hand and Inspire Hand.} The first two rows correspond to the LEAP Hand, and the third row corresponds to the Inspire Hand. Each object is evaluated over 10 trials, and the success rate is reported in parentheses.}
  \vspace{-15pt}
  \label{fig:real_grasp}
\end{figure}

\textbf{LEAP Hand Results.} This HOI dataset includes eight object categories: \emph{Wine Glass}, \emph{Tong}, \emph{Syringe}, \emph{Phone}, \emph{Power Drill}, \emph{Microphone}, \emph{Spray Bottle}, \emph{Mug}. 
In real-world experiments, our approach achieves an average success rate of 83.75\,\% across eight objects (Fig.~\ref{fig:real_grasp}). It performs well on objects that support form-closure functional grasps, such as the \emph{Power Drill} and \emph{Microphone}. It also produces functional grasps on the \emph{Spray Bottle} and \emph{Mug}, placing fingers on the spray head and handle. 

\textbf{Inspire Hand Results.} The Inspire Hand training dataset includes four object categories: \emph{Bowl}, \emph{Plate}, \emph{Knife}, and \emph{Fork}, with success rates of 10/10, 8/10, 3/10, and 5/10, respectively. The Inspire Hand features soft finger layers that increase both friction and contact area, allowing it to securely grasp small objects such as the \emph{Fork}. However, the \emph{Fork} and \emph{Knife} remain challenging due to their small size; grasp attempts often miss by a small margin because of pose estimation inaccuracies. Across the two dexterous hands, we achieve an overall success rate of 77.5\% on 12 objects, using models trained solely on web images.

\section{Conclusion, Limitations, and Future Work}
\label{sec:conclusion}
This work proposes a method for learning functional grasps from 3D reconstructed hand–object interaction (HOI) data collected from web images, eliminating the need for human-collected grasp demonstrations. Experimental results demonstrate that the proposed approach can effectively leverage reconstructed HOI data to train grasping models that generate human-like functional grasps.

Despite its advantages, Web2Grasp has several limitations. 
(1) The reconstruction process can fail for certain object categories (Fig.~\ref{fig:grasp_comparison}). Since our method is agnostic to the choice of mesh reconstruction and can benefit from improvements in reconstruction techniques, future work will explore more advanced image-to-mesh methods, such as Trellis~\cite{xiang2024structured}, to improve reconstruction quality.
(2) Current functional grasps tend to follow a similar pattern for each object, whereas some objects may support multiple grasps depending on the task. Incorporating task descriptions via language conditioning could enable the model to generate diverse, task-specific grasps.
(3) We currently filter out failed reconstructed grasps using geometric metrics. Future work will investigate optimization-based methods, such as ContactOpt~\cite{grady2021contactopt}, to optimize these grasps and make them useful for training.

\bibliographystyle{IEEEtran}
\bibliography{IEEEabrv, example}

\end{document}